# Data-Driven Approximation of Binary-State Network Reliability Function: Algorithm Selection and Reliability Thresholds for Large-Scale Systems

Wei-Chang Yeh
Department of Industrial Engineering and Engineering Management
National Tsing Hua University
Hsinchu, Taiwan, R.O.C.

*Abstract:* Network reliability assessment is pivotal for ensuring the robustness of modern infrastructure systems, from power grids to communication networks. While exact reliability computation for binary-state networks is NP-hard/#P-hard, existing approximation methods face critical tradeoffs between accuracy, scalability, and data efficiency. This study evaluates 20 machine learning methods across three reliability regimes—full range (0.0–1.0), high reliability (0.9–1.0), and ultra-high reliability (0.99–1.0)—to address these gaps. We demonstrate that large-scale networks with arc reliability ≥0.9 exhibit near-unity system reliability, enabling computational simplifications. Further, we establish a dataset-scale-driven paradigm for algorithm selection: Artificial Neural Networks (ANN) excel with limited data ($size < m^2$), while Polynomial Regression (PR) achieves superior accuracy in data-rich environments ($size \geq m^2$). Our findings reveal ANN's Test-MSE of 7.24E−05 at 30,000 samples and PR's optimal performance (5.61E−05) at 40,000 samples, outperforming traditional Monte Carlo simulations. These insights provide actionable guidelines for balancing accuracy, interpretability, and computational efficiency in reliability engineering, with implications for infrastructure resilience and system optimization.

## 1. INTRODUCTION

Modern infrastructure systems—from power grids and communication networks to IoT ecosystems—demand rigorous reliability analysis to ensure operational resilience. These systems are often modeled as binary-state networks, where components (arcs/nodes) operate in either functional (1) or failed (0) states [1, 2, 3]. Within this paradigm, network reliability—the probability of maintaining connectivity between specified nodes under given conditions—serves as a critical performance metric [4, 5–7].

Existing methods for binary-state network reliability analysis fall into two broad categories: exact and approximated. Exact approaches, such as Binary Addition Tree (BAT) algorithms [2, 8–9] and Binary Decision Diagrams [10], derive precise reliability values but face NP-hard/#P-hard complexity, rendering them impractical for large-scale networks [2, 10–11]. Approximated methods, like Monte Carlo Simulation (MCS) [12–14], prioritize scalability through probabilistic sampling but trade accuracy for efficiency, particularly in rare-event scenarios [15]. To reconcile these tradeoffs, Yeh et al. introduced BAT-MCS [14], a hybrid framework combining deterministic BAT principles with stochastic MCS. This method leverages a self-regulating simulation mechanism to reduce variance and enhance accuracy, addressing the limitations of standalone exact or approximated techniques.

Despite these advances, critical challenges persist. First, exact methods remain computationally prohibitive for large or dynamically evolving networks. Second, MCS demands extensive simulations to stabilize estimates, straining resources in data-scarce environments. Third, emerging machine learning (ML) and artificial intelligence (AI) techniques—though promising—lack systematic evaluation for reliability approximation under limited data constraints. This gap is particularly pressing, as real-world applications (e.g., aging infrastructure, sensor networks) often lack sufficient failure data to train data-hungry models.

This study addresses these challenges through a comprehensive evaluation of 20 machine learning methods across three reliability regimes: full range (0.0–1.0), high reliability (0.9–1.0), and ultra-high reliability (0.99–1.0). We introduce two key contributions:

1. Component Reliability Threshold: Demonstrating that networks with arc reliability ≥0.9 exhibit near-unity system reliability, enabling computational simplifications for large-scale systems.
2. Dataset-Scale-Driven Algorithm Selection: Establishing Artificial Neural Networks (ANN) as optimal for small datasets ($size < m^2$), while Polynomial Regression (PR) excels with large datasets ($size \geq m^2$), where $m$ denotes the polynomial degree.

The remainder of this paper is organized as follows: Section 2 defines key notations, acronyms, and assumptions foundational to our analysis. Section 3 details the Binary Addition Tree (BAT) algorithm, its integration with Monte Carlo Simulation (BAT-MCS), and introduces the 20 machine learning methods evaluated in this study. Section 4 presents experimental results across three reliability regimes—full range (0.0–1.0), high reliability (0.9–1.0), and ultra-high reliability (0.99–1.0)—analyzing performance trends, scalability, and the impact of dataset size on algorithm accuracy. Finally, Section 5 synthesizes key findings, including component reliability thresholds and dataset-driven algorithm selection criteria, discusses limitations, and proposes future research directions for dynamic and multi-state network reliability assessment.

## 2. CONCEPTUAL FRAMEWORK: TERMS, SYMBOLS, AND ASSUMPTIONS

This section outlines the essential acronyms, mathematical notations, key assumptions, and nomenclature required to contextualize and implement the proposed GNN-BAT-MCS methodology.

### 2.1 Acronyms

BAT: Binary-Addition-Tree Algorithm (efficient combinatorial search for network states)

MCS: Monte Carlo simulation (probabilistic sampling for reliability estimation)

BAT-MCS BAT-integrated MCS with self-adaptive simulation-number tuning

LSA: Layered Search Algorithm

PLSA: Path-based LSA (LSA variant for path-centric reliability evaluation)

### 2.2 Notations

$|\bullet|$ : number of elements in a set

$||\bullet||$ : number of coordinates in a vector/subvector

$a_i$ : arc $i$ in the network

$V$ : node set where $V = \{1, 2, \ldots, n\}$

$E$ : arc set where $E = \{a_1, a_2, \ldots, a_m\}$

$X$ : Binary state vector

$X(a_i)$ : value of arc $a_i$ in state vector $X$ (e.g., for X = (1, 1, 1, 0, 0), $X(a_1) = X(a_2) = X(a_3) = 1$ and $X(a_4) = X(a_5) = 0$).

Pr(•) : probability of an event. For example, Pr($X$) represents the probability of event $X$ occurring. If $X$ is a state vector, Pr($X$) gives the probability of the system being in that state.

**D** : Arc state distribution listed $\mathbf{D}(a) = \Pr(a)$ for all $a \in E$. example shown in **Table 1**.

**Table 1.** State distribution **D**.

| $i$ | $\mathbf{D}(a_i)$ |
|---|---|
| 1 | 0.90 |
| 2 | 0.80 |
| 3 | 0.70 |
| 4 | 0.60 |
| 5 | 0.50 |

$G(V, E)$ : Undirected graph with node set $V$ and arc set $E$. Example, **Figure 1** shows a graph where: $V = \{1, 2, 3, 4\}$, $E = \{a_1, a_2, a_3, a_4, a_5\}$, node 1 is source, and node 4 is target.

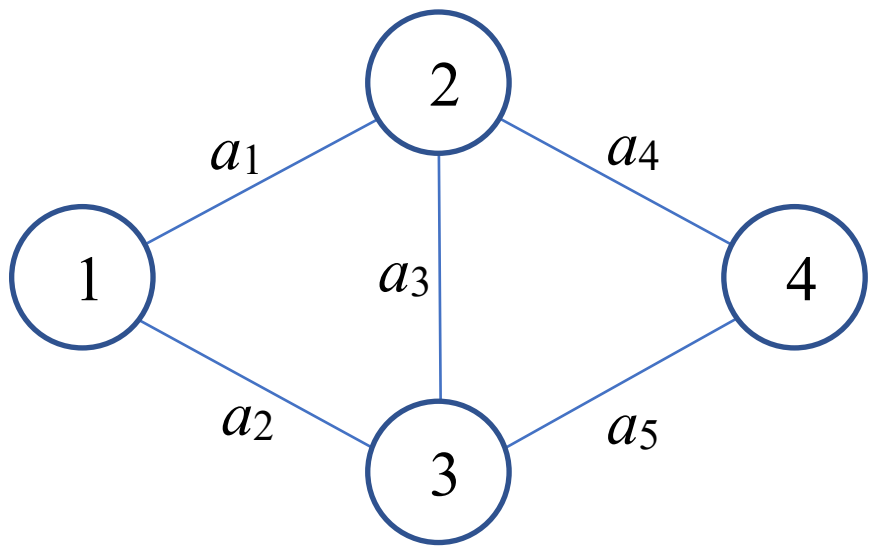

**Figure 1.** Example graph.

$G(V, E, \mathbf{D})$ : Binary-state network with structure $G(V, E)$ and distribution **D** (Example: **Figure 1** with distribution from **Table 1**).

$G(X)$ : Subgraph corresponding to state vector $X$, where $G(X) = G(V, \{a \in E \mid X(a) = 1\}, \mathbf{D})$. Example: For $X = (1, 1, 1, 0, 0)$, $G(X)$ shows working arcs as in **Figure 2**.

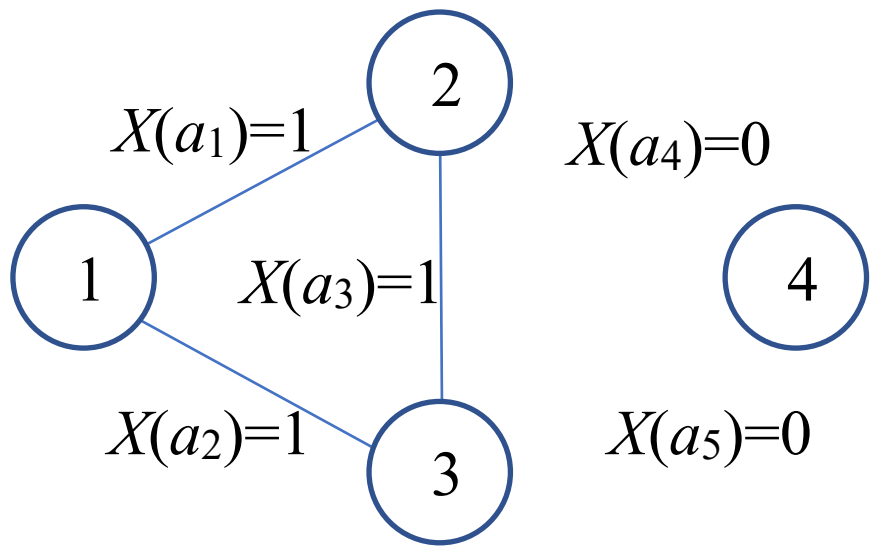

**Figure 2.** $X = (1, 1, 1, 0, 0)$ and $G(X)$, where $G(V, E)$ is shown in Figure 1.

$R$ : exact reliability

$R_{\text{BAT-MCS}}(\bullet)$ : approximated reliability obtained from the BAT-MCS

$R_{\text{GM-MCS}}(\bullet)$ : predicted approximated reliability from the GM-MCS

$\delta$ : number of coordinates in supervectors

$N_{sim}$ : Number of simulations per MCS/BAT-MCS run

$N_{sim}(S)$ : number of simulations of supervector $S$ in each BAT-MCS run

$N_{run}$ : number of runs for each BAT-MCS

$\varepsilon$ : error between the estimator and the exact solution

$\sigma$ : standard deviation

$\epsilon$ : the threshold that determines the difference between the observed values and the values predicted by the model.

$w$: the model parameters (coefficients)

$y_i$: the observed target values

$\hat{y}_i$: the predicted values for $y_i$

$l(y_i, \hat{y}_i)$: the individual loss function, e.g., Mean Squared Error (MSE) and Binary Cross-Entropy, quantifies the error between $y_i$ and $\hat{y}_i$

$X_i$: the feature vectors

$N$: the number of samples

$\lambda$: the regularization parameter

$||w||_2^2$ the L2 norm (sum of squared values) of the parameter vector

$||w||_1$ the L1 norm (sum of absolute values) of the parameter vector

**2.3 Nomenclature**

Binary-state network: A network where each arc exists in one of two states: functional (1) or failed (0).

Reliability: The probability that the network successfully connects the source node (node 1) and target node (node $n$).

Supervector: A vector $S = (S(a_1), S(a_2), ..., S(a_\delta))$ derived from binary state $X$, where $S(a_i) = X(a_i)$ for $i = 1, 2, \ldots, \delta$ and $\delta \leq m$ ($m$= total arcs).

Connected vector: A state vector $X$ where the network graph $G(\mathbf{X})$ contains at least one operational path between the source and target nodes.

Disconnected vector: A state vector $X$ where no operational path exists between the source and target nodes in $G(X)$.

### 2.4 Assumptions

The reliability analysis in this study is based on the following fundamental assumptions:

1. Network Structure:
   - No parallel arcs exist between any two nodes
   - No self-loops are present in the network
2. Node Properties:
   - All nodes maintain 100% reliability
   - All nodes are interconnected within the network
3. Arc Characteristics:
   - Each arc operates in a binary state (working/failed)
   - Arc states are statistically independent of one another

## 3. REVIEW OF BAT-MCS, BAT, MCS, PLSA, AND 20 MACHINE LEARNING METHODS

This section describes the BAT-MCS framework [14], combining three key methodologies (Binary Addition Tree (BAT), Monte Carlo Simulation (MCS), and Path-Based Layered Search Algorithm (PLSA)) for reliability approximation. It also introduces the 20 machine learning methods compared in this study.

### 3.1 BAT-MCS: Hybrid Integration for Scalable Reliability Estimation

The Binary Addition Tree Monte Carlo Simulation (BAT-MCS) algorithm integrates the complementary strengths of two established approaches: the systematic combinatorial analysis of Binary Addition Tree (BAT) and the efficient random sampling of Monte Carlo Simulation (MCS) [14]. This hybrid method leverages BAT's ability to precisely handle combinatorial aspects while exploiting MCS's computational efficiency for large-scale systems. This subsection provides a detailed examination of the BAT and MCS algorithms individually, followed by an analysis of their integration in BAT-MCS.

### 3.1.1 Binary Addition Tree (BAT)

The Binary Addition Tree (BAT) algorithm, introduced by Yeh [16], systematically enumerates all $2^m$ $m$-tuple binary-state vectors without duplicates using iterative application of two fundamental rules [16]:

1. Saturation Rule: Locate the first failed arc (denoted as $a$) in $X$. Reset the states of all arcs preceding $a$ to zero and change the state of $a$ to working to form a new state vector.
2. Termination Rule: If no failed arc exists, the algorithm terminates.

Its computational complexity is $O(2^m)$, making it highly efficient for combinatorial state generation. The pseudocode is structured as follows:

**Algorithm: BAT**

**Input:** Positive integer $m$ (number of arcs).

**Output:** All binary-state vectors $X$.

**Initialize:** Coordinate index $i = 1$, vector $X = (0, 0, \ldots, 0)$.

**Iterate:**

**While** $i \leq m$:

**If** $X(a_i)$: Set $X(a_i) = 1$, reset $I = 1$, record $X$, repeat.

**Else**: Set $X(a_i) = 0$, increment $i = i+1$, repeat.

BAT's simplicity stems from its three-step structure, requiring only vector updates and no auxiliary data structures. It generates all $2^m$ vectors in linear space complexity $O(m)$, outperforming traditional methods like DFS, BFS, and UGFM in runtime and flexibility [17–19]. Applications span network reliability [20–23], IoT systems [24, 25], supply chain resilience [11, 26–27], traveling salesman problem [28], resilience problems [29], computer virus problems [30], redundancy allocation problem [31], and heterogeneous-arc network reliability problems [32].

### 3.1.2 Monte Carlo Simulation (MCS)

Monte Carlo Simulation (MCS) approximates network reliability by statistically sampling arc states. For $N_{sim}$ trials, MCS generates random numbers $\rho_i \sim U[0, 1]$ for each arc $a_i$, sets $X(a_i) = 1$ (operational) if $\rho_i < \Pr(a_i)$, and checks connectivity via PLSA. The unbiased reliability estimator $R_{MCS}$ is:

$$R_{MCS} = N_{pass}/N_{sim} \tag{1}$$

where $N_{pass}$ is the count of connected vectors.

**Algorithm: MCS**

**Input:** Network $G(V, E, \mathbf{D})$ and $N_{sim}$.

**Output:** Approximated reliability $R_{MCS}$.

**Initialize:** $N_{pass} = 0$.

**Iterate:**

For sim = 1 to $N_{sim}$:

Generate $X$ by sampling $\rho_i$ for all $a_i \in E$.

If $X$ is connected: Increment $N_{pass}$.

**Return:** $R_{MCS} = N_{pass}/N_{sim}$.

The relative error $\epsilon$ and confidence interval $(1-\alpha)$% are derived from:

$$\frac{Z_{\frac{\alpha}{2}}^{2}(1-R_{MCS})}{\varepsilon^2 R_{MCS}} \leq N_{sim}. \tag{2}$$

### 3.1.3 BAT-MCS Example and Concept

By focusing on **supervectors**—partial state vectors with δ coordinates (δ ≪ $m$)—BAT-MCS reduces computational complexity while maintaining accuracy. Each supervector $S$ deterministically fixes the states of δ arcs, while MCS probabilistically samples the remaining $m-\delta$ arcs. This hybrid approach minimizes variance and computational overhead, particularly in large-scale networks.

**Algorithm: BAT-MCS**

**Input:** Network $G(V, E, \mathbf{D})$, source node 1, target node $n$, supervector dimension δ, and simulation count $N_{sim}$.

**Output:** Approximated reliability $R_{BAT\text{-}MCS}$.

**Initialize:** $R_{\text{BAT-MCS}} = 0$.

**Generate Supervectors:** Use BAT to enumerate all δ-tuple supervectors $S \in \Omega$.

**Prune Redundant Supervectors:**

- Connected Supervectors: For each $S$, compute Zeor($S$) = ($S(a_1)$, …, $S(a_\delta)$, 0, …, 0). If Zeor($S$) is connected (via PLSA), add Pr($S$) to $R_{\text{BAT-MCS}}$.
- Disconnected Supervectors: For One($S$) = ($S(a_1)$, …, $S(a_\delta)$, 1, …, 1), if One($S$) is disconnected, discard $S$.

**Adaptive MCS for Ambiguous Supervectors:**

- For remaining $S \in \Omega$, compute:

$$\mathrm{N_{sim}}(S) = \left\lfloor \mathrm{N_{sim}} \times \frac{\Pr(S)}{\sum_{S\in\Omega}\Pr(S)} \right\rfloor. \tag{3}$$

- Perform MCS with $\mathrm{N_{sim}}(S)$ trials for each $S$, updating:

$$R_{\text{BAT-MCS}} = R_{\text{BAT-MCS}} + \Pr(S) \times \frac{\mathrm{N_{pass}}(S)}{\mathrm{N_{sim}}(S)}. \tag{4}$$

The time complexity of BAT-MCS is dominated by $O(2^{\delta}\times(m-\delta)\times\mathrm{N_{sim}}(S))$, significantly faster than standalone BAT ($O(2^m)$) when $\delta \ll m$. From the computational experiments, the BAT-MCS outperforms the MCS in terms of solution quality and variance [14].

For example, consider a network (Figure 1) with $m$ = 5 arcs, $\mathrm{N_{sim}}$ = 980, and δ = 2, i.e., ($a_1$, $a_2$). BAT generates four supervectors. After pruning, $\Omega = \{S_2 = (1, 0), S_3 = (0, 1), S_4 = (1, 1)\}$, Pr($S_2$) = Pr($S_2(a_1)$) × Pr($S_2(a_2)$) = 0.9 × 0.2 = 0.18, Pr($S_3$) = 0.08, Pr($S_4$) = 0.72, $\mathrm{N_{sim}}(S_2)$ = 180, $\mathrm{N_{sim}}(S_3)$ = 80, and $\mathrm{N_{sim}}(S_4)$ = 720.

For each $S_i \in \Omega$, MCS samples the remaining arcs $a_3$, $a_4$, and $a_5$ (Table 4). Connectivity is verified via PLSA. Asume that we have $\mathrm{N_{pass}}(S_2)$ = 105, $\mathrm{N_{pass}}(S_2)$ = 75, and $\mathrm{N_{pass}}(S_2)$ = 454. Hence, $R_{\text{BAT-MCS}}(S_2)$ = Pr($S_2$) × $\mathrm{N_{pass}}(S_2)/\mathrm{N_{sim}}(S_2)$ = 0.1050, $\mathrm{N_{pass}}(S_2)$ = 0.0750, $\mathrm{N_{pass}}(S_2)$ = 0.4540, and $R_{\text{BAT-MCS}}$ = 0.1050 + 0.0750 + 0.4540 = 0.634.

### 3.2 Path-Based Layered Search Algorithm (PLSA)

Network reliability calculation requires verifying connectivity between source (node 1) and target (node *n*) for each state vector *X*. The Path-Based Layered Search Algorithm (PLSA) [14], an extension of the Layered Search Algorithm (LSA) [33], efficiently determines connectivity in $O(n)$ time, where $n$ is the number of nodes.

**Algorithm: PLSA**

**Input: Binary-state vector *X*.**

**Output:** Connectivity status of *X*.

**Initialize:** Layer $L_1 = \{1\}$, layer index $i = 2$, and $L_2 = \varnothing$.

**Iterate:**

Construct $L_i$ as nodes reachable from $L_{i-1}$ via operational arcs in *X*.

**If** $L_i = \varnothing$: *X* is disconnected.

**If** $n \in L_i$: *X* is connected.

**Else**: Increment *I* and repeat.

**Example Application:** For $X = (1, 1, 1, 1, 1)$ in **Figure 1**, identifies connectivity in 3 layers (**Table 2**).

**Table 2.** Example PLSA procedure on $X = (1, 1, 1, 1, 1)$ in Figure 1.

| Layer $i$ | Nodes in $L_i$ | Connectivity Status |
|---|---|---|
| 1 | {1} | |
| 2 | {2, 3} | |
| 3 | {4} | Connected |

### 3.3 20 Machine Learning Methods

This study evaluates 20 distinct machine learning approaches, encompassing traditional regression and sophisticated ensemble techniques. The methods examined include:

- Regression Models: Huber Regressor, Polynomial Regression, Bayesian Ridge Regression, Ridge Regression, Linear Regression, ElasticNet, Lasso, Kernel Ridge Regression.
- Tree-Based & Ensemble Methods: XGBoost, LightGBM, CatBoost, Gradient Boosting,

Random Forest, Extra Trees, Decision Tree, AdaBoost.

- Support Vector Methods: SVR (Linear Kernel), SVR (RBF Kernel).
- Neural Networks: Neural Network (Artificial Neural Network, ANN),
- Instance-Based Learning: K-Nearest Neighbors (KNN).

The following sections provide a concise theoretical overview and essential mathematical formulations for each algorithm.

### 3.3.1 Regression Models

(1) Linear Regression

The simplest regression model that fits a linear relationship between independent variables and the target and it is defined as:

$$y = Xw + \epsilon \tag{5}$$

The objective is to minimize the sum of squared errors, which can be expressed mathematically as:

$$\min_{w} \sum_{i=1}^{N} (y_i - X_i w)^2 \tag{6}$$

(2) Ridge Regression

Ridge Regression is a linear regression model that incorporates L2 regularization to prevent overfitting. With a regularization parameter of λ = 1.0 used in this study, Ridge Regression is defined as:

$$\min_{w} \sum_{i=1}^{N} (y_i - X_i w)^2 + \lambda \left\| w \right\|_2^2. \tag{7}$$

(3) Lasso Regression

Lasso Regression is a regression model that employs L1 regularization to enforce sparsity in the solution, effectively setting some coefficients to zero (with λ = 0.1). It is defined as:

$$\min_{w} \sum_{i=1}^{N} (y_i - X_i w)^2 + \lambda \left\| w \right\|_1. \tag{8}$$

(4) ElasticNet Regression

A linear regression model with both L1 (Lasso) and L2 (Ridge) regularization ($\lambda_1 = 0.5, \lambda_2 = 0.1$). It is defined as:

$$\min_{w} \sum_{i=1}^{N}(y_i - X_i w)^2 + \lambda_1 ||w||_1 + \lambda_2 ||w||_2^2 \,. \tag{9}$$

(5) Polynomial Regression

An extension of linear regression where the model includes polynomial terms of the input features (degree = 2). For a multi-variable polynomial regression of degree 2, the predicted output $\hat{y}$ is given by:

$$\hat{y} = w_0 + \sum_{i=1}^{m} w_i x_i + \sum_{i=1}^{m} \sum_{j=i}^{m} w_{ij} x_i x_j \tag{10}$$

where:

- $x_1, x_2, \dots, x_m$ are the input variables.
- $w_0$ is the intercept (bias term).
- $w_i$ and $w_{ij}$ are the model coefficients (weights).
- The terms include higher-order interactions up to degree 2.

(6) Bayesian Ridge Regression

Bayesian Ridge Regression is a Bayesian approach to ridge regression that introduces a prior on the coefficients to prevent overfitting. It can be expressed as:

$$p(w|X, y) \propto p(y|X, w)p(w) \tag{11}$$

(7) Huber Regressor

A robust regression method that minimizes the Huber loss, which is a combination of squared loss (for small errors) and absolute loss (for large errors). It is less sensitive to outliers and its complete loss function with regularization is defined as:

$$\min_{w} \sum_{i=1}^{N} H_\epsilon(y_i - X_i w) + \lambda ||w||_2^2 \tag{12}$$

where:

- $$H_\epsilon(y, \hat{y}) = \begin{cases} \frac{1}{2}(y - \hat{y})^2 & if|y - \hat{y}| \le \epsilon \\ \epsilon\left(|y - \hat{y}| - \frac{\epsilon}{2}\right) & \text{otherwise} \end{cases} \tag{13}$$
- $\epsilon = 1.35$ is the threshold for treating residuals as outliers, which is commonly used to balance robustness and efficiency.
- $\lambda = 0.0001$ is the regularization parameter, controlling the penalty on model complexity.

(8) Kernel Ridge Regression

Kernel Ridge Regression is a combination of ridge regression and the kernel trick to model non-linearity. It can be expressed as:

$$\min_{f \in H} \sum_{i=1}^{N} (y_i - f(X_i))^2 + \lambda \left|\left|f\right|\right|_H^2 \tag{14}$$

where $f$ is a function in the reproducing kernel Hilbert space H associated with a kernel $K$ and $\left|\left|f\right|\right|_H^2$ is the norm of $f$ in the Hilbert space.

### 3.3.2 Tree-Based & Ensemble Methods

(1) Decision Tree

A tree-based model that recursively splits data into subsets by selecting features that best separate classes or minimize variance. For example, the Gini Index serves as a criterion for splitting nodes in classification tasks, quantifying node impurity. It is defined as:

$$G = 1 - \sum_{i=1}^{c} p_i^2, \tag{15}$$

where $p_i$ is the probability of class $i$ in the node.

(2) Random Forest

An ensemble of decision trees where each tree is trained on a random subset of data and features. The final prediction is obtained by averaging the outputs of all trees, expressed as:

$$\hat{y} = \frac{1}{T} \sum_{t=1}^{T} h_t(x), \tag{16}$$

where $h_t(x)$ represents the prediction of the $t$-th tree, and $T$ is the total number of trees in the ensemble.

(3) Extra Trees (Extremely Randomized Trees)

An ensemble method like Random Forest but selects split points randomly to reduce variance. This increased randomness helps to further reduce variance and improve generalization.

(4) Gradient Boosting

An ensemble method that builds models sequentially, each correcting errors from the previous one. Sequential ensemble method minimizing loss function gradient:

$$F_m(x) = F_{m-1}(x) + \gamma h_m(x), \tag{17}$$

where $F_m(x)$ represents the model's prediction at iteration $m$, $h_m(x)$ is trained to approximate the negative gradient of loss function, and $\gamma$ is a learning rate that controls the contribution of each new model.

(5) AdaBoost (Adaptive Boosting)

A boosting algorithm that assigns higher weights to misclassified instances in each iteration. Boosts weak learners by focusing on misclassified samples:

$$w_i^{(t+1)} = w_i^{(t)} e^{-\alpha_t y_i h_t(x_i)} \tag{18}$$

where $w_i^{(t)}$ represents the weight of sample $i$ at iteration $t$, $h_t(x_i)$ is the prediction of the weak learner, $y_i$ is the true label, and $\alpha_t$ is a scaling factor that controls the contribution of each weak learner.

(6) XGBoost (Extreme Gradient Boosting)

A powerful gradient boosting algorithm optimized for speed and performance using regularization and tree pruning techniques. Optimized gradient boosting with regularization:

$$L(\theta) = \sum_{i=1}^{N} l(y_i, \hat{y_i}) + \sum_{k=1}^{K} \Omega(f_k) \tag{19}$$

where $\Omega(f) = \gamma T + \frac{1}{2}\lambda||w||^2$ is the regularization term for trees, $T$ is the number of leaves, $w$ denotes the leaf weights, and $\gamma$ and $\lambda$ are regularization parameters that help control model complexity and prevent overfitting.

(7) LightGBM

A gradient boosting framework that uses decision trees for fast and efficient learning, optimized for large datasets. The loss function is defined as:

$$L = \sum_{i=1}^{N} l(y_i, \hat{y_i}) + \lambda||w||^2 \tag{20}$$

(8) CatBoost

A gradient boosting library optimized for categorical features, improving model accuracy and training efficiency. It employs an innovative method called ordered boosting to reduce overfitting and bias when dealing with categorical data. The loss function for CatBoost is defined as:

$$L = \sum_{i=1}^{N} l(y_i, \hat{y_i}) + \Omega(T), \tag{21}$$

where $\Omega(T)$ is the regularization term that controls the complexity of the decision trees, helping to prevent overfitting.

### 3.3.3 Support Vector Methods

(1) SVR (Support Vector Regression) - Linear Kernel

A regression version of SVM that tries to fit the best margin within a given threshold ($\epsilon$) by using a linear function $\hat{y} = w^T x + b$ as the regression function. To learn $w$ and $b$, SVR minimizes the following objective function:

$$\min_{w,b,\xi_i,\xi_i^*} \frac{1}{2}||w||^2 + C\sum_{i=1}^{m}(\xi_i + \xi_i^*) \tag{22}$$

subject to:

$$y_i - (w^T x_i + b) \leq \epsilon + \xi_i$$

$$(w^T x_i + b) - y_i \leq \epsilon + \xi_i^*$$

$$\xi_i, \xi_i^* \geq 0, i = 1,2, \dots, m \tag{23}$$

where:

- $\epsilon = 0.1$ is the epsilon-insensitive margin, meaning errors within $\epsilon$ are ignored,
- $\xi_i, \xi_i^*$ are slack variables that allow some errors beyond $\epsilon$,
- $C$ =100 is the regularization parameter, controlling the trade-off between model complexity and tolerance to misclassification.

(2) SVR (RBF Kernel)

A non-linear SVR that uses the Radial Basis Function (RBF) kernel to model complex relationships. SVR with an RBF kernel solves the following objective function:

$$\min_{\alpha,\alpha^*} \frac{1}{2}\sum_{i,j}(\alpha_i - \alpha_i^*)(\alpha_j - \alpha_j^*)K(x_i, x_j) + \epsilon\sum_i(\alpha_i + \alpha_i^*) - \sum_i y_i(\alpha_i - \alpha_i^*) \tag{24}$$

subject to:

$$0 \leq \alpha_i, \alpha_i^* \leq C,$$

$$\sum_i(\alpha_i - \alpha_i^*) = 0, \tag{25}$$

where:

- $\alpha_i, \alpha_i^*$ are Lagrange multipliers learned from the optimization problem,
- the RBF kernel function $K(x_i, x_j) = \exp(-\gamma|x_i - x_j|^2)$, $\gamma = 0.1$ is the kernel coefficient that controls how much influence a single training sample has,

- $C = 100$ is the regularization parameter that controls the trade-off between smoothness and fitting the training data,
- $\epsilon = 0.1$ is the epsilon-insensitive margin, meaning predictions within $\pm\epsilon$ of the true value are not penalized.
- $\gamma$ = 'scale': Automatically sets $\gamma = \frac{1}{n \cdot \sigma^2}$, where $n$ is the number of features and $\sigma^2$ is the variance of the dataset.

**3.3.4 Neural Network**

(1) Neural Network (ANN - Artificial Neural Network)

The proposed regression model employs a multi-layer perceptron (MLP) with backpropagation, leveraging deep learning principles to approximate complex non-linear relationships. The network comprises an input layer, hidden layers, and an output layer, with each neuron processing weighted inputs through non-linear activation functions. Details of the framework are as follows:

- Forward pass: For layer $l$, the activation $a^{(l)}$ s computed as:

$$a^{(l)} = \sigma\left(W^{(l)} a^{(l-1)} + b^{(l)}\right), \tag{26}$$

  where $\sigma$ is the ReLU activation function, $W^{(l)}$ denotes layer weights, and $b^{(l)}$ is the bias vector.
- Backpropagation: $\frac{\partial J}{\partial W^{(l)}} = \delta^{(l+1)}\left(a^{(l)}\right)^T$, where $\delta^{(l)}$ is the backpropagated error.
- Weight update via Adam optimizer: $W^{(l)} = W^{(l)} - \alpha \frac{\partial J}{\partial W^{(l)}}$.

The model employs the Adam optimizer for adaptive weight updates, balancing convergence speed and stability through momentum-guided gradient adjustments. A two-layer architecture (10 neurons in the first hidden layer, 5 in the second) ensures task-specific complexity without overparameterization. Training is restricted to 1,000 epochs, with early stopping triggered if validation loss plateaus for 10 consecutive epochs, enforcing robust generalization.

**3.3.5 Instance-Based Learning**

(1) K-Nearest Neighbors (KNN)

A non-parametric model that predicts based on the average of the K-nearest training points. In this study, $K = 5$.

$$y = \frac{1}{K}\sum_{i=1}^{K} y_i. \tag{27}$$

## 4 EXPERIMENTAL COMPUTATIONS

To evaluate the performance of the 20 machine learning algorithms, two computational experiments (Ex1 and Ex2) were conducted on binary-state networks.

- Ex1: Focused on benchmark networks ranging from small to medium sizes.
- Ex2: Validated findings on a larger-scale network.

### 4.1 Shared Experimental Setup

Calculating exact reliability for binary-state networks is both NP-hard and #P-hard, even with simplified component states. To ensure equitable comparisons, all reliability values were estimated using BAT-MCS [14] with $N_{sim}$ = 10, 000 simulations. The BAT-MCS implementation was executed in C++ (VS Code) on a 64-bit Windows 10 platform equipped with an Intel Core i7-6650U processor (2.20 GHz) and 16 GB RAM.

All 20 machine learning methods**,** Huber Regressor, Polynomial Regression, Bayesian Ridge Regression, Ridge Regression, Linear Regression, ElasticNet, Lasso, Kernel Ridge Regression, XGBoost, LightGBM, CatBoost, Gradient Boosting, Random Forest, Extra Trees, Decision Tree, AdaBoost, SVR (Linear Kernel), SVR (RBF Kernel), Artificial Neural Network, and KNN, were implemented in Python under identical hardware conditions. The dataset, derived from arc reliability approximations, was partitioned into training (80%) and testing (20%) subsets. To ensure robustness and mitigate overfitting, 5-fold cross-validation was employed during model training.

To evaluate the effectiveness of different methods for approximating the reliability function, we conducted a series of computational experiments. The experiments assessed the accuracy of 20 algorithms across 20 benchmark networks under three distinct arc reliability ranges: 0.0–1.0 (full range), 0.9–1.0 (high-reliability), and 0.99–1.0 (ultra-high-reliability). Specifically, we divided the experiments into three parts based on the range of reliability values:

1. All-Range Reliability (0.0–1.0): This experiment considers the entire range of possible reliability values, providing a comprehensive evaluation of how well the methods approximate reliability across different system conditions, from complete failure to full functionality. This assessment helps determine the overall robustness of each approach.

2. Higher Reliability (0.9–1.0): Many practical systems operate under high-reliability conditions, where reliability is expected to be close to 1. This experiment focuses on cases where reliability is at least 0.9, testing the ability of the methods to provide accurate approximations in scenarios where failures are rare but still critical to analyze.
3. Ultra-High-Reliability (0.99–1.00): Some highly dependable systems, such as safety-critical infrastructure or advanced manufacturing processes, require extremely high reliability. This experiment examines the performance of the methods in the narrow range of 0.99 to 1.00, where small estimation errors can have significant implications.

### 4.2 Experiment 1 (Ex 1)

Experiment 1 evaluates 20 different methods on small to medium-sized benchmark networks to determine which performs best across three reliability ranges: [0.0, 1.0], [0.9, 1.0], and [0.99, 1.0]. Each dataset contains $N_{run}$ = 10,000 samples. This yielded a comprehensive dataset of 1.2 million experimental instances (20 networks × 3 ranges × 20 algorithms × 10,000 samples).

#### 4.2.1 Test Networks

Figure 1 showcases 20 established binary-state network benchmarks widely used in network reliability research. These benchmarks exhibit diverse topologies, scales, and application contexts, providing a comprehensive framework for evaluating algorithmic performance across various scenarios. The networks serve as standard validation tools for assessing new algorithmic approaches in network reliability studies [16, 34–36] and as reference points for validating approximation accuracy. This diversity highlights the inherent trade-offs between computational tractability and topological complexity in reliability analysis. Table 3 provides node and arc counts for each benchmark network.

It's important to note that computational complexity remains significant even for modest network structures. For example, the network shown in Figure 7(1), despite having only 30 arcs, presents substantial computational challenges for reliability calculation even when limited to a single fixed time step. This underscores the inherent complexity of network reliability analysis and emphasizes the value of efficient approximation methods.

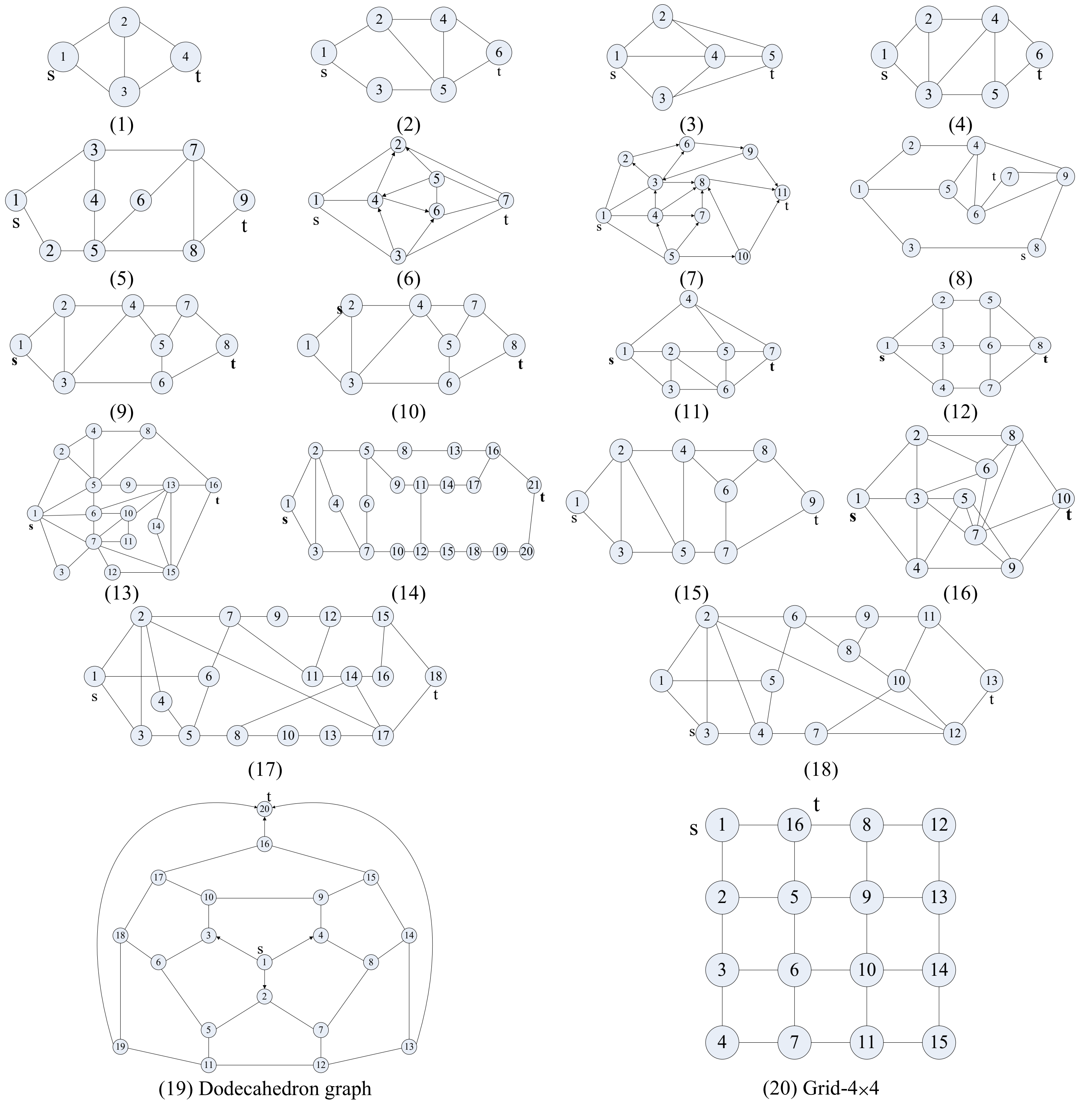

**Figure** 3. 20 benchmark binary-state networks used in the test

**Table 3.** Node number *n* and arc number *m* of each benchmark network.

| Network | Nodes (*n*) | Arcs (*m*) | Network | Nodes (*n*) | Arcs (*m*) |
|---|---|---|---|---|---|
| Fig. 3 (1) | 4 | 5 | Fig. 3 (11) | 7 | 12 |
| Fig. 3 (2) | 6 | 8 | Fig. 3 (12) | 8 | 13 |
| Fig. 3 (3) | 5 | 8 | Fig. 3 (13) | 16 | 30 |
| Fig. 3 (4) | 6 | 9 | Fig. 3 (14) | 21 | 29 |
| Fig. 3 (5) | 9 | 12 | Fig. 3 (15) | 9 | 14 |
| Fig. 3 (6) | 7 | 14 | Fig. 3 (16) | 10 | 21 |
| Fig. 3 (7) | 11 | 21 | Fig. 3 (17) | 18 | 27 |
| Fig. 3 (8) | 9 | 13 | Fig. 3 (18) | 13 | 22 |
| Fig. 3 (9) | 8 | 12 | Fig. 3 (19) | 20 | 30 |
| Fig. 3 (10) | 8 | 12 | Fig. 3 (20) | 16 | 24 |

#### 4.2.2 The Observations

Performance was quantified using four metrics: Train MSE, Test MSE, Test MAE, and Cross-Validation (CV) Score. Key results are distilled in Tables 4–6 due to space limitations. Table 4 highlights the top three algorithms per reliability range based on Test MSE. Table 5 details statistical summaries of approximated reliability instances in Experiment 1 (Ex1), including the number of instances with approximated reliability equal to one ($N_1$), mean (Avg), standard deviation (Stdev), median (Med), minimum (Min), maximum (Max), and range (Range), while Table 6 provides a full ranking of all algorithms across networks and reliability ranges.

The observations of Experiment 1 (Ex 1) are summarized in the following based on three distinct arc reliability ranges:

1. Full Reliability Range (0.0–1.0):

This case exhibited the highest complexity, as reflected in the elevated Test MSE and Test MAE values compared to narrower reliability ranges (Tables 4 and 5). The inclusion of networks spanning both extremely low and high reliability necessitated adaptability to diverse behavioral patterns, complicating approximation.

In Table 6, Polynomial Regression (PR) demonstrated superior performance with an average ranking of 1.3 across 20 problem instances, significantly outperforming other methods. XGBoost (XGB, rank: 3.8) and LightGBM (LGBM, rank: 4.0) followed, while ElasticNet (EN, rank: 19.8), Lasso (LA, rank: 19.0), and Decision Tree (DT, rank: 16.6) consistently underperformed.

2. High-Reliability Range (0.9–1.0):

The reduced variability in this range facilitated improved model performance (Tables 4 and 5). As in Table 6, PR again dominated with an average rank of 3.0, surpassing CatBoost (CB, rank: 3.5) and LightGBM (LGBM, rank: 3.7). Conversely, Kernel Ridge (KR, rank: 20.0), Artificial Neural Networks (ANN, rank: 18.3), and SVR with a linear kernel (SVR-L, rank: 18.1) delivered the weakest results.

3. Ultra-High-Reliability Range (0.99–1.0):

Networks in this range exhibited enhanced robustness due to structural redundancy, with an average reliability of 0.999915 (Table 4). Only 15 of 20 benchmark instances (ID < 16) yielded valid results, as the remaining exceeded computational precision thresholds. For solvable cases, all methods achieved exceptional accuracy, with MAE values below $9.51 \times 10^{-8}$.

**Table 4.** Results for the top three algorithms tested on 20 benchmark networks in Ex1.

| ID | Algo. | Train MSE | Test MSE | Test MAE | CV Score | Algo. | Train MSE | Test MSE | Test MAE | CV Score | Algo. | Train MSE | Test MSE | Test MAE | CV Score |
|---|---|---|---|---|---|---|---|---|---|---|---|---|---|---|---|
| 1 | KR | 0.000152 | 0.000263 | 0.011144 | 0.000259 | PR | 7.5E-07 | 7.39E-07 | 0.000669 | 7.86E-07 | PR | 7.52E-09 | 6.75E-09 | 6.33E-05 | 7.67E-09 |
| | PR | 0.000930 | 0.001181 | 0.022049 | 0.001058 | XGB | 5.02E-07 | 1.08E-06 | 0.000799 | 1.18E-06 | BR | 7.81E-09 | 6.83E-09 | 6.34E-05 | 7.65E-09 |
| | ET | 0.000000 | 0.001187 | 0.024410 | 0.001162 | LGBM | 5.39E-07 | 1.1E-06 | 0.000809 | 1.23E-06 | LR | 7.81E-09 | 6.84E-09 | 6.35E-05 | 7.65E-09 |
| 2 | KR | 0.000186 | 0.000690 | 0.016153 | 0.000723 | PR | 1.16E-06 | 1.1E-06 | 0.000832 | 1.28E-06 | LR | 1.33E-08 | 1.18E-08 | 7.99E-05 | 1.33E-08 |
| | PR | 0.001292 | 0.001408 | 0.026700 | 0.001490 | XGB | 8.31E-07 | 2.35E-06 | 0.001192 | 2.36E-06 | BR | 1.33E-08 | 1.18E-08 | 7.98E-05 | 1.33E-08 |
| | ET | 0.000000 | 0.002018 | 0.034089 | 0.002854 | CB | 1.64E-06 | 2.39E-06 | 0.001246 | 2.33E-06 | HR | 1.35E-08 | 1.19E-08 | 7.84E-05 | 1.36E-08 |
| 3 | PR | 0.001407 | 0.001291 | 0.027854 | 0.001626 | PR | 6.54E-08 | 5.75E-08 | 0.000182 | 7E-08 | BR | 2.45E-11 | 4.97E-11 | 9.08E-07 | 2.99E-11 |
| | KR | 0.000262 | 0.001505 | 0.023994 | 0.001073 | ET | 0 | 7.06E-08 | 0.000195 | 8.78E-08 | RR | 2.49E-11 | 4.98E-11 | 7.28E-07 | 2.99E-11 |
| | XGB | 0.001031 | 0.002873 | 0.039535 | 0.003483 | LGBM | 3.72E-08 | 7.07E-08 | 0.000191 | 8.24E-08 | LA | 2.49E-11 | 4.98E-11 | 7.36E-07 | 2.99E-11 |
| 4 | KR | 0.000239 | 0.001434 | 0.022923 | 0.001560 | PR | 4.78E-07 | 5.18E-07 | 0.000558 | 2.13E-06 | HR | 5.57E-09 | 4.36E-09 | 4.87E-05 | 5.52E-09 |
| | PR | 0.001440 | 0.001674 | 0.029995 | 0.001749 | CB | 5.27E-07 | 7.15E-07 | 0.000682 | 7.95E-07 | BR | 5.33E-09 | 4.46E-09 | 5.16E-05 | 5.26E-09 |
| | ANN | 0.003079 | 0.004185 | 0.050739 | 0.005026 | XGB | 3.2E-07 | 7.78E-07 | 0.000688 | 7.96E-07 | LR | 5.32E-09 | 4.5E-09 | 5.14E-05 | 5.26E-09 |
| 5 | PR | 0.001098 | 0.001281 | 0.026724 | 0.001488 | PR | 7.58E-07 | 1.03E-06 | 0.000785 | 0.000001 | BR | 8.08E-09 | 7.53E-09 | 6.72E-05 | 8.19E-09 |
| | KR | 0.000152 | 0.002236 | 0.028607 | 0.002573 | CB | 1.08E-06 | 1.84E-06 | 0.001046 | 1.74E-06 | CB | 6.42E-09 | 7.59E-09 | 6.67E-05 | 8.37E-09 |
| | XGB | 0.001158 | 0.003694 | 0.046177 | 0.003791 | XGB | 5.59E-07 | 1.9E-06 | 0.00106 | 1.9E-06 | LR | 8.07E-09 | 7.62E-09 | 6.73E-05 | 8.2E-09 |
| 6 | PR | 0.001229 | 0.001636 | 0.029319 | 0.001602 | PR | 9.11E-08 | 8.76E-08 | 0.000224 | 1.04E-07 | BR | 8.55E-11 | 9.85E-11 | 1.94E-06 | 8.93E-11 |
| | KR | 0.000301 | 0.002360 | 0.032840 | 0.002364 | CB | 8.81E-08 | 1.26E-07 | 0.000265 | 1.36E-07 | LR | 8.51E-11 | 9.89E-11 | 2.21E-06 | 8.99E-11 |
| | LGBM | 0.000890 | 0.002748 | 0.039584 | 0.002962 | LGBM | 5.7E-08 | 1.32E-07 | 0.000268 | 1.44E-07 | RR | 8.68E-11 | 9.9E-11 | 1.88E-06 | 8.93E-11 |
| 7 | PR | 0.001508 | 0.002401 | 0.034617 | 0.002509 | PR | 7.11E-08 | 9.62E-08 | 0.00024 | 9.56E-08 | XGB | 7.45E-11 | 4.3E-13 | 6.56E-07 | 6.02E-11 |
| | LGBM | 0.001792 | 0.006308 | 0.062475 | 0.006744 | LGBM | 4.19E-08 | 1.07E-07 | 0.00024 | 1.03E-07 | RR | 7.45E-11 | 6E-13 | 7.75E-07 | 6.02E-11 |
| | XGB | 0.001741 | 0.006465 | 0.062793 | 0.006982 | GB | 3.36E-08 | 1.1E-07 | 0.000241 | 1.09E-07 | LA | 7.45E-11 | 6E-13 | 7.75E-07 | 6.02E-11 |
| 8 | PR | 0.000013 | 0.000018 | 0.003265 | 0.000017 | BR | 9.08E-06 | 8.7E-06 | 0.00234 | 9.28E-06 | HR | 5.81E-09 | 3.99E-09 | 4.75E-05 | 5.71E-09 |
| | ET | 0.000000 | 0.000071 | 0.006099 | 0.000077 | LR | 9.08E-06 | 8.71E-06 | 0.00234 | 9.28E-06 | BR | 5.58E-09 | 4.26E-09 | 5.2E-05 | 5.45E-09 |
| | GB | 0.000064 | 0.000235 | 0.011832 | 0.000234 | HR | 9.19E-06 | 8.9E-06 | 0.002357 | 9.31E-06 | LR | 5.56E-09 | 4.29E-09 | 5.16E-05 | 5.45E-09 |
| 9 | PR | 0.001579 | 0.001852 | 0.032420 | 0.002235 | PR | 4.92E-07 | 8.12E-07 | 0.000687 | 6.68E-07 | ET | 0 | 4.73E-09 | 5.36E-05 | 5.21E-09 |
| | KR | 0.000211 | 0.002492 | 0.031491 | 0.003505 | GB | 3.08E-07 | 1.13E-06 | 0.000838 | 1.14E-06 | RF | 7.42E-10 | 4.95E-09 | 5.48E-05 | 5.31E-09 |
| | XGB | 0.001426 | 0.004857 | 0.052675 | 0.004829 | XGB | 3.81E-07 | 1.18E-06 | 0.00085 | 1.05E-06 | CB | 3.77E-09 | 4.96E-09 | 5.42E-05 | 5.19E-09 |
| 10 | PR | 0.001420 | 0.001972 | 0.033265 | 0.002009 | PR | 6.5E-07 | 7.34E-07 | 0.000676 | 8.96E-07 | CB | 3.65E-09 | 5.92E-09 | 5.9E-05 | 5.21E-09 |
| | KR | 0.000144 | 0.004166 | 0.043500 | 0.004059 | CB | 8.63E-07 | 1.05E-06 | 0.000798 | 1.3E-06 | ET | 0 | 5.96E-09 | 6.02E-05 | 5.32E-09 |
| | XGB | 0.001396 | 0.005709 | 0.056397 | 0.005338 | XGB | 4.72E-07 | 1.16E-06 | 0.000852 | 1.33E-06 | BR | 4.69E-09 | 5.99E-09 | 5.98E-05 | 5.1E-09 |
| 11 | PR | 0.001287 | 0.002035 | 0.033052 | 0.001793 | PR | 2.48E-08 | 2.6E-08 | 0.000119 | 3.01E-08 | RR | 4.98E-11 | 9.93E-11 | 1.47E-06 | 5.98E-11 |
| | GB | 0.001013 | 0.003526 | 0.046190 | 0.004915 | GB | 1.17E-08 | 2.89E-08 | 0.000128 | 3.4E-08 | LA | 4.98E-11 | 9.93E-11 | 1.47E-06 | 5.98E-11 |
| | XGB | 0.001329 | 0.003608 | 0.045819 | 0.004609 | LGBM | 1.42E-08 | 2.98E-08 | 0.000131 | 3.32E-08 | EN | 4.98E-11 | 9.93E-11 | 1.47E-06 | 5.98E-11 |
| 12 | PR | 0.001483 | 0.001852 | 0.032997 | 0.002028 | PR | 1.07E-07 | 1.44E-07 | 0.000297 | 1.44E-07 | LA | 4.98E-11 | 4.98E-11 | 9.72E-07 | 4.99E-11 |
| | KR | 0.000213 | 0.003522 | 0.041592 | 0.004651 | CB | 1.21E-07 | 1.69E-07 | 0.000319 | 1.86E-07 | EN | 4.98E-11 | 4.98E-11 | 9.72E-07 | 4.99E-11 |
| | LGBM | 0.001499 | 0.004481 | 0.053382 | 0.005193 | LGBM | 6.98E-08 | 1.74E-07 | 0.000329 | 1.86E-07 | XGB | 4.98E-11 | 4.98E-11 | 9.13E-07 | 4.99E-11 |
| 13 | PR | 0.001039 | 0.002111 | 0.035786 | 0.002360 | LGBM | 1.43E-08 | 2.45E-08 | 0.000123 | 3.21E-08 | RR | 2.49E-11 | 4.98E-11 | 7.36E-07 | 3E-11 |
| | LGBM | 0.001607 | 0.005610 | 0.058746 | 0.006640 | GB | 1.24E-08 | 2.6E-08 | 0.000125 | 3.36E-08 | LA | 2.49E-11 | 4.98E-11 | 7.36E-07 | 3E-11 |
| | XGB | 0.001699 | 0.005796 | 0.059749 | 0.006669 | CB | 2.23E-08 | 2.63E-08 | 0.000128 | 3.22E-08 | EN | 2.49E-11 | 4.98E-11 | 7.36E-07 | 3E-11 |
| 14 | ET | 0.000000 | 0.000951 | 0.013948 | 0.000442 | PR | 3.54E-06 | 1.28E-05 | 0.002804 | 1.35E-05 | BR | 8.63E-08 | 9.51E-08 | 0.000244 | 9.32E-08 |
| | PR | 0.000130 | 0.000951 | 0.018503 | 0.000644 | HR | 0.000014 | 1.78E-05 | 0.003271 | 1.58E-05 | LR | 8.62E-08 | 9.55E-08 | 0.000245 | 9.33E-08 |
| | LGBM | 0.000164 | 0.001231 | 0.015908 | 0.000655 | BR | 1.38E-05 | 1.84E-05 | 0.003272 | 1.55E-05 | HR | 8.66E-08 | 9.62E-08 | 0.000246 | 9.35E-08 |
| 15 | PR | 0.001073 | 0.001283 | 0.026243 | 0.001525 | PR | 1.36E-06 | 2.08E-06 | 0.001135 | 1.91E-06 | BR | 1.69E-08 | 1.76E-08 | 0.000104 | 1.76E-08 |
| | KR | 0.000099 | 0.002754 | 0.032664 | 0.002742 | XGB | 1.35E-06 | 3.51E-06 | 0.001429 | 4.22E-06 | LR | 1.69E-08 | 1.76E-08 | 0.000104 | 1.76E-08 |
| | GB | 0.000814 | 0.003193 | 0.041005 | 0.003702 | LGBM | 1.49E-06 | 4.06E-06 | 0.001597 | 4.75E-06 | HR | 1.71E-08 | 1.77E-08 | 0.000103 | 1.78E-08 |
| 16 | PR | 0.000868 | 0.001716 | 0.032030 | 0.002026 | PR | 2.09E-08 | 4.17E-08 | 0.000151 | 0.578191 | | | | | |
| | XGB | 0.001130 | 0.004276 | 0.048366 | 0.004214 | LGBM | 1.42E-08 | 4.56E-08 | 0.000154 | 0.538538 | | | | | |
| | GB | 0.000924 | 0.004315 | 0.048891 | 0.004329 | GB | 1.3E-08 | 4.59E-08 | 0.000157 | 0.535317 | | | | | |
| 17 | PR | 0.000534 | 0.002602 | 0.040667 | 0.002464 | LGBM | 3.4E-07 | 1.14E-06 | 0.000823 | 0.932889 | | | | | |
| | XGB | 0.000820 | 0.003428 | 0.043814 | 0.003560 | XGB | 3.1E-07 | 1.23E-06 | 0.000845 | 0.927515 | | | | | |
| | GB | 0.000726 | 0.003529 | 0.045672 | 0.003574 | CB | 7.35E-07 | 1.24E-06 | 0.000829 | 0.926921 | | | | | |
| 18 | PR | 0.001037 | 0.002205 | 0.034850 | 0.002643 | ET | 0 | 3.4E-07 | 0.000427 | 0.936935 | | | | | |
| | LGBM | 0.001489 | 0.005313 | 0.058142 | 0.006023 | CB | 2.23E-07 | 3.56E-07 | 0.000446 | 0.934017 | | | | | |
| | GB | 0.001267 | 0.005551 | 0.058057 | 0.006196 | LGBM | 1.24E-07 | 3.71E-07 | 0.000445 | 0.931319 | | | | | |
| 19 | HR | 0.005523 | 0.003973 | 0.046890 | 0.005534 | LR | 2.98E-07 | 2.56E-07 | 0.000387 | 0.75582 | | | | | |
| | PR | 0.000482 | 0.003996 | 0.050418 | 0.003687 | BR | 2.98E-07 | 2.56E-07 | 0.000387 | 0.755387 | | | | | |
| | XGB | 0.001312 | 0.004054 | 0.046105 | 0.005374 | HR | 3.05E-07 | 2.6E-07 | 0.000383 | 0.751615 | | | | | |
| 20 | PR | 0.000379 | 0.001133 | 0.024788 | 0.001165 | CB | 7.32E-07 | 8.71E-07 | 0.000715 | 0.95907 | | | | | |
| | LGBM | 0.000391 | 0.001162 | 0.020893 | 0.001116 | GB | 3.45E-07 | 9.05E-07 | 0.000728 | 0.957456 | | | | | |
| | XGB | 0.000347 | 0.001281 | 0.023656 | 0.001262 | ET | 0 | 9.75E-07 | 0.000751 | 0.954134 | | | | | |

**Table 5.** Statistical summary of approximated reliability instances in Experiment 1.

| ID | (0.0-1.0)-reliability | | | | | | | (0.9-1.0)-reliability | | | | | | | (0.99-1.0)-reliability | | | | | | |
|---|---|---|---|---|---|---|---|---|---|---|---|---|---|---|---|---|---|---|---|---|---|
| | $N_1$ | Avg. | Stdev. | Min | Max | Range | Med | $N_1$ | Avg. | Stdev. | Min | Max | Range | Med | $N_1$ | Avg. | Stdev. | Min | Max | Range | Med |
| 1 | 0 | 0.463919 | 0.240276 | 0.0134 | 0.9862 | 0.9728 | 0.4527 | 1 | 0.992546 | 0.00456 | 0.9753 | 1 | 0.0247 | 0.9933 | 511 | 0.999924 | 0.000098 | 0.9995 | 1 | 0.0005 | 1 |
| 2 | 0 | 0.31523 | 0.204811 | 0.0006 | 0.9141 | 0.9135 | 0.2784 | 0 | 0.987985 | 0.006599 | 0.9654 | 0.9998 | 0.0344 | 0.9886 | 376 | 0.99988 | 0.000131 | 0.9991 | 1 | 0.0009 | 0.9999 |
| 3 | 0 | 0.621528 | 0.219848 | 0.0699 | 0.9946 | 0.9247 | 0.6445 | 90 | 0.999378 | 0.000565 | 0.9961 | 1 | 0.0039 | 0.9995 | 996 | 1 | 0.000005 | 0.9999 | 1 | 0.0001 | 1 |
| 4 | 0 | 0.372729 | 0.208294 | 0.0058 | 0.9332 | 0.9274 | 0.3526 | 0 | 0.994505 | 0.003271 | 0.9804 | 0.9999 | 0.0195 | 0.995 | 639 | 0.999949 | 0.000079 | 0.9996 | 1 | 0.0004 | 1 |
| 5 | 0 | 0.249045 | 0.182268 | 0.0014 | 0.9742 | 0.9728 | 0.2067 | 0 | 0.99132 | 0.004784 | 0.9741 | 0.9996 | 0.0255 | 0.9919 | 496 | 0.999922 | 0.000097 | 0.9994 | 1 | 0.0006 | 0.9999 |
| 6 | 0 | 0.551648 | 0.226436 | 0.0264 | 0.9925 | 0.9661 | 0.5537 | 51 | 0.999039 | 0.000788 | 0.9956 | 1 | 0.0044 | 0.9992 | 990 | 0.999999 | 0.000009 | 0.9999 | 1 | 0.0001 | 1 |
| 7 | 0 | 0.415493 | 0.210637 | 0.0206 | 0.9534 | 0.9328 | 0.4016 | 65 | 0.999207 | 0.00066 | 0.9959 | 1 | 0.0041 | 0.9994 | 993 | 0.999999 | 0.000008 | 0.9999 | 1 | 0.0001 | 1 |
| 8 | 0 | 0.259656 | 0.223707 | 0 | 0.9781 | 0.9781 | 0.1957 | 0 | 0.902528 | 0.039638 | 0.8073 | 0.9977 | 0.1904 | 0.9013 | 615 | 0.999946 | 0.000081 | 0.9995 | 1 | 0.0005 | 1 |
| 9 | 0 | 0.34921 | 0.207454 | 0.0118 | 0.9511 | 0.9393 | 0.3163 | 0 | 0.993959 | 0.003505 | 0.9823 | 0.9998 | 0.0175 | 0.9945 | 630 | 0.99995 | 0.000076 | 0.9996 | 1 | 0.0004 | 1 |
| 10 | 0 | 0.25325 | 0.174988 | 0.0038 | 0.9036 | 0.8998 | 0.2199 | 0 | 0.993619 | 0.003535 | 0.978 | 0.9997 | 0.0217 | 0.994 | 630 | 0.99995 | 0.000076 | 0.9996 | 1 | 0.0004 | 1 |
| 11 | 0 | 0.537777 | 0.22167 | 0.0228 | 0.9964 | 0.9736 | 0.546 | 243 | 0.999744 | 0.000269 | 0.9981 | 1 | 0.0019 | 0.9998 | 993 | 0.999999 | 0.000008 | 0.9999 | 1 | 0.0001 | 1 |
| 12 | 0 | 0.422582 | 0.208217 | 0.006 | 0.9746 | 0.9686 | 0.4153 | 15 | 0.998761 | 0.000876 | 0.9941 | 1 | 0.0059 | 0.9989 | 994 | 0.999999 | 0.000007 | 0.9999 | 1 | 0.0001 | 1 |
| 13 | 0 | 0.576066 | 0.199897 | 0.0716 | 0.9929 | 0.9213 | 0.5865 | 193 | 0.999716 | 0.000269 | 0.9982 | 1 | 0.0018 | 0.9998 | 996 | 1 | 0.000005 | 0.9999 | 1 | 0.0001 | 1 |
| 14 | 0 | 0.020246 | 0.040225 | 0 | 0.4482 | 0.4482 | 0.0069 | 0 | 0.927738 | 0.020913 | 0.8529 | 0.9744 | 0.1215 | 0.9293 | 1 | 0.999112 | 0.000405 | 0.9975 | 1 | 0.0025 | 0.9991 |
| 15 | 0 | 0.19653 | 0.161757 | 0.0009 | 0.81 | 0.8091 | 0.1535 | 0 | 0.983248 | 0.008458 | 0.9557 | 0.9988 | 0.0431 | 0.9838 | 236 | 0.999822 | 0.000158 | 0.9991 | 1 | 0.0009 | 0.9998 |
| 16 | 0 | 0.70953 | 0.18512 | 0.0584 | 0.9979 | 0.9395 | 0.7383 | 247 | 0.999735 | 0.000284 | 0.9982 | 1 | 0.0018 | 0.9998 | 999 | 1 | 0 | 1 | 1 | 0 | 1 |
| 17 | 0 | 0.206826 | 0.16905 | 0.0001 | 0.8116 | 0.8115 | 0.1586 | 1 | 0.993665 | 0.003983 | 0.979 | 1 | 0.021 | 0.9942 | 640 | 0.999951 | 0.000077 | 0.9996 | 1 | 0.0004 | 1 |
| 18 | 0 | 0.455563 | 0.216179 | 0.0169 | 0.9841 | 0.9672 | 0.453 | 27 | 0.99732 | 0.002324 | 0.9891 | 1 | 0.0109 | 0.998 | 812 | 0.999978 | 0.00005 | 0.9997 | 1 | 0.0003 | 1 |
| 19 | 0 | 0.166247 | 0.122406 | 0.0025 | 0.6452 | 0.6427 | 0.1374 | 6 | 0.99825 | 0.00113 | 0.9929 | 1 | 0.0071 | 0.9984 | 983 | 0.999998 | 0.000013 | 0.9999 | 1 | 0.0001 | 1 |
| 20 | 0 | 0.565772 | 0.265885 | 0.0046 | 0.9974 | 0.9928 | 0.5677 | 4 | 0.993059 | 0.004779 | 0.9767 | 1 | 0.0233 | 0.9938 | 526 | 0.999925 | 0.000099 | 0.9995 | 1 | 0.0005 | 1 |
| Avg | 0 | 0.385442 | 0.194456 | 0.01688 | 0.9120 | 0.89509 | 0.36927 | 47 | 0.987266 | 0.00556 | 0.96927 | 0.99849 | 0.02922 | 0.98763 | 703 | 0.999915 | 7.41E-05 | 0.99955 | 1 | 0.00045 | 0.9994 |

**Table 6.** Algorithm rankings across network topologies and reliability ranges.

| Algo. | (0.0-1.0)-reliability | | | | | | | | | | | | | | | | | | | | |
|---|---|---|---|---|---|---|---|---|---|---|---|---|---|---|---|---|---|---|---|---|---|
| | 1 | 2 | 3 | 4 | 5 | 6 | 7 | 8 | 9 | 10 | 11 | 12 | 13 | 14 | 15 | 16 | 17 | 18 | 19 | 20 | Avg |
| AB | 12 | 17 | 15 | 17 | 16 | 17 | 17 | 8 | 17 | 17 | 16 | 16 | 14 | 13 | 16 | 14 | 11 | 13 | 15 | 13 | 14.7 |
| BR | 14 | 14 | 13 | 14 | 13 | 14 | 9 | 15 | 12 | 8 | 12 | 12 | 7 | 12 | 12 | 8 | 9 | 10 | 4 | 9 | 11.1 |
| CB | 10 | 9 | 10 | 10 | 9 | 8 | 12 | 6 | 9 | 13 | 6 | 7 | 10 | 5 | 7 | 5 | 5 | 5 | 8 | 6 | 8.0 |
| DT | 18 | 18 | 18 | 18 | 18 | 18 | 18 | 9 | 18 | 18 | 18 | 18 | 17 | 11 | 18 | 15 | 15 | 15 | 18 | 15 | 16.6 |
| EN | 20 | 20 | 20 | 20 | 20 | 20 | 20 | 20 | 20 | 20 | 20 | 20 | 20 | 19 | 20 | 19 | 20 | 20 | 20 | 18 | 19.8 |
| ET | 3 | 3 | 4 | 8 | 5 | 7 | 13 | 2 | 5 | 6 | 8 | 8 | 11 | **1** | 5 | 6 | 6 | 6 | 12 | 5 | 6.2 |
| GB | 5 | 6 | 5 | 6 | 7 | 4 | 4 | 3 | 6 | 4 | 2 | 4 | 4 | 6 | 3 | 3 | 3 | 3 | 9 | 4 | 4.6 |
| HR | 16 | 12 | 16 | 16 | 15 | 12 | 7 | 13 | 10 | 11 | 15 | 10 | 6 | 15 | 15 | 11 | 13 | 8 | **1** | 8 | 11.5 |
| KR | **1** | **1** | 2 | **1** | 2 | 2 | 5 | 11 | 2 | 2 | 5 | 2 | 18 | 14 | 2 | 20 | 17 | 18 | 17 | 19 | 8.1 |
| KNN | 11 | 10 | 11 | 11 | 17 | 11 | 16 | 18 | 16 | 16 | 17 | 17 | 16 | 8 | 17 | 17 | 16 | 16 | 13 | 17 | 14.6 |
| LA | 19 | 19 | 19 | 19 | 19 | 19 | 19 | 19 | 19 | 19 | 19 | 19 | 19 | 18 | 19 | 18 | 19 | 19 | 19 | 20 | 19.0 |
| LGBM | 4 | 4 | 6 | 5 | 4 | 3 | 2 | 5 | 4 | 5 | 4 | 3 | 2 | 3 | 6 | 4 | 4 | 2 | 7 | 2 | 4.0 |
| LR | 15 | 13 | 14 | 15 | 12 | 15 | 8 | 14 | 13 | 10 | 13 | 11 | 8 | 9 | 11 | 9 | 10 | 11 | 6 | 10 | 11.4 |
| ANN | 9 | 11 | 7 | 3 | 6 | 6 | 6 | 10 | 15 | 15 | 10 | 15 | 12 | 16 | 9 | 13 | 14 | 17 | 11 | 12 | 10.9 |
| PR | 2 | 2 | **1** | 2 | **1** | **1** | **1** | **1** | **1** | **1** | **1** | **1** | **1** | 2 | **1** | **1** | **1** | **1** | 2 | **1** | **1.3** |
| RF | 6 | 7 | 9 | 9 | 8 | 10 | 15 | 4 | 8 | 14 | 9 | 9 | 13 | 7 | 8 | 7 | 7 | 7 | 14 | 7 | 8.9 |
| RR | 13 | 15 | 12 | 12 | 14 | 13 | 10 | 17 | 11 | 9 | 11 | 13 | 5 | 10 | 13 | 10 | 8 | 9 | 5 | 11 | 11.1 |
| SVR-L | 17 | 16 | 17 | 13 | 11 | 16 | 11 | 12 | 14 | 7 | 14 | 14 | 9 | 17 | 10 | 12 | 12 | 12 | 10 | 14 | 12.9 |
| SVR-R | 8 | 8 | 8 | 7 | 10 | 9 | 14 | 16 | 7 | 12 | 7 | 6 | 15 | 20 | 14 | 16 | 18 | 14 | 16 | 16 | 12.1 |
| XGB | 7 | 5 | 3 | 4 | 3 | 5 | 3 | 7 | 3 | 3 | 3 | 5 | 3 | 4 | 4 | 2 | 2 | 4 | 3 | 3 | 3.8 |

| Algo. | (0.9-1.0)-reliability | | | | | | | | | | | | | | | | | | | | |
|---|---|---|---|---|---|---|---|---|---|---|---|---|---|---|---|---|---|---|---|---|---|
| | 1 | 2 | 3 | 4 | 5 | 6 | 7 | 8 | 9 | 10 | 11 | 12 | 13 | 14 | 15 | 16 | 17 | 18 | 19 | 20 | Avg |
| AB | 12 | 11 | 12 | 11 | 11 | 12 | 11 | 11 | 11 | 11 | 12 | 11 | 10 | 12 | 11 | 10 | 8 | 8 | 10 | 8 | 10.7 |
| BR | 10 | 8 | 9 | 8 | 8 | 9 | 9 | **1** | 10 | 8 | 8 | 7 | 6 | 3 | 8 | 8 | 10 | 11 | 2 | 10 | 7.7 |
| CB | 4 | 3 | 4 | 2 | 2 | 2 | 4 | 6 | 5 | 2 | 5 | 2 | 3 | 5 | 4 | 4 | 3 | 2 | 6 | **1** | 3.5 |
| DT | 13 | 13 | 13 | 13 | 13 | 13 | 12 | 12 | 12 | 13 | 14 | 14 | 14 | 14 | 13 | 14 | 12 | 9 | 14 | 12 | 12.9 |
| EN | 16 | 16 | 16 | 16 | 16 | 16 | 16 | 19 | 16 | 16 | 16 | 16 | 16 | 17 | 16 | 16 | 16 | 16 | 16 | 16 | 16.2 |
| ET | 5 | 6 | 2 | 6 | 6 | 6 | 6 | 5 | 6 | 6 | 4 | 9 | 4 | 9 | 6 | 5 | 5 | **1** | 8 | 3 | 5.4 |
| GB | 6 | 5 | 5 | 4 | 4 | 5 | 3 | 9 | 2 | 5 | 2 | 5 | 2 | 8 | 5 | 3 | 4 | 6 | 7 | 2 | 4.6 |
| HR | 9 | 10 | 10 | 10 | 9 | 11 | 10 | 3 | 8 | 9 | 11 | 6 | 7 | 2 | 7 | 9 | 11 | 12 | 3 | 9 | 8.3 |
| KR | 20 | 20 | 20 | 20 | 20 | 20 | 20 | 20 | 20 | 20 | 20 | 20 | 20 | 20 | 20 | 20 | 20 | 20 | 20 | 20 | 20.0 |
| KNN | 8 | 12 | 11 | 12 | 12 | 8 | 13 | 14 | 13 | 12 | 7 | 12 | 13 | 10 | 12 | 12 | 14 | 13 | 13 | 14 | 11.8 |
| LA | 15 | 15 | 15 | 15 | 15 | 15 | 15 | 18 | 15 | 15 | 15 | 15 | 15 | 16 | 15 | 15 | 15 | 15 | 15 | 15 | 15.2 |
| LGBM | 3 | 4 | 3 | 5 | 5 | 3 | 2 | 10 | 4 | 4 | 3 | 3 | **1** | 7 | 3 | 2 | **1** | 3 | 4 | 4 | 3.7 |
| LR | 11 | 9 | 8 | 9 | 7 | 10 | 8 | 2 | 9 | 7 | 9 | 8 | 8 | 4 | 9 | 7 | 9 | 10 | **1** | 11 | 7.8 |
| ANN | 17 | 17 | 19 | 19 | 19 | 19 | 19 | 13 | 19 | 19 | 19 | 19 | 19 | 15 | 19 | 19 | 19 | 19 | 19 | 19 | 18.3 |
| PR | **1** | **1** | **1** | **1** | **1** | **1** | **1** | 4 | **1** | **1** | **1** | **1** | 11 | **1** | **1** | **1** | 7 | 7 | 11 | 6 | **3.0** |
| RF | 7 | 7 | 6 | 7 | 10 | 7 | 7 | 8 | 7 | 10 | 6 | 10 | 5 | 11 | 10 | 6 | 6 | 5 | 9 | 7 | 7.6 |
| RR | 14 | 14 | 14 | 14 | 14 | 14 | 14 | 15 | 14 | 14 | 13 | 13 | 12 | 13 | 14 | 13 | 13 | 14 | 12 | 13 | 13.6 |
| SVR-L | 19 | 18 | 18 | 18 | 18 | 18 | 18 | 17 | 18 | 18 | 18 | 18 | 18 | 19 | 18 | 18 | 18 | 18 | 18 | 18 | 18.1 |
| SVR-R | 18 | 19 | 17 | 17 | 17 | 17 | 17 | 16 | 17 | 17 | 17 | 17 | 17 | 18 | 17 | 17 | 17 | 17 | 17 | 17 | 17.2 |
| XGB | 2 | 2 | 7 | 3 | 3 | 4 | 5 | 7 | 3 | 3 | 10 | 4 | 9 | 6 | 2 | 11 | 2 | 4 | 5 | 5 | 4.9 |

| Algo. | (0.99-1.0)-reliability | | | | | | | | | | | | | | | |
|---|---|---|---|---|---|---|---|---|---|---|---|---|---|---|---|---|
| | 1 | 2 | 3 | 4 | 5 | 6 | 7 | 8 | 9 | 10 | 11 | 12 | 13 | 14 | 15 | Avg |
| AB | 14 | 16 | 16 | 14 | 16 | 15 | 15 | 7 | 6 | 5 | 7 | 9 | 5 | 7 | 9 | 11.2 |
| BR | 2 | **1** | **1** | 5 | 2 | 4 | 3 | 5 | 7 | 6 | **1** | **1** | 6 | **1** | **1** | **2.9** |
| CB | 5 | 2 | 10 | 8 | 4 | 3 | **1** | 11 | 14 | 13 | 4 | 4 | 13 | 4 | 4 | 6.8 |
| DT | 15 | 14 | 12 | 16 | 14 | 16 | 16 | 16 | 5 | 15 | 12 | 15 | 15 | 12 | 15 | 13.7 |
| EN | 12 | 12 | 5 | 4 | 11 | 13 | 11 | 3 | 2 | 3 | 15 | 14 | 3 | 15 | 14 | 9.3 |
| ET | 4 | 5 | 13 | 7 | 7 | **1** | 2 | 9 | 10 | 8 | 10 | 6 | 8 | 10 | 6 | 7.4 |
| GB | 6 | 7 | 8 | 11 | 5 | 6 | 12 | 15 | 13 | 14 | 8 | 7 | 14 | 8 | 7 | 9.2 |
| HR | **1** | 4 | 7 | 15 | **1** | 9 | 6 | 12 | 9 | 16 | 3 | 3 | 16 | 3 | 3 | 6.9 |
| KR | 20 | 20 | 20 | 19 | 20 | 20 | 20 | 20 | 19 | 19 | 20 | 20 | 19 | 20 | 20 | 19.7 |
| KNN | 9 | 15 | 15 | 9 | 13 | 10 | 14 | 10 | 15 | 11 | 11 | 11 | 11 | 11 | 11 | 11.7 |
| LA | 11 | 11 | 4 | 3 | 10 | 12 | 10 | 2 | **1** | 2 | 14 | 13 | 2 | 14 | 13 | 8.3 |
| LGBM | 7 | 6 | 11 | 10 | 6 | 7 | 5 | 8 | 11 | 9 | 5 | 5 | 9 | 5 | 5 | 7.8 |
| LR | 3 | 3 | 2 | 6 | 3 | 5 | 4 | 6 | 8 | 7 | 2 | 2 | 7 | 2 | 2 | 4.1 |
| ANN | 19 | 19 | 19 | 20 | 19 | 19 | 19 | 19 | 20 | 20 | 19 | 19 | 20 | 19 | 19 | 19.3 |
| PR | 16 | 8 | 9 | 13 | 15 | 8 | 13 | 14 | 16 | 12 | 16 | 16 | 12 | 16 | 16 | 11.7 |
| RF | 8 | 9 | 14 | 12 | 8 | 2 | 7 | 13 | 12 | 10 | 9 | 8 | 10 | 9 | 8 | 9.3 |
| RR | 10 | 10 | 3 | 2 | 9 | 11 | 8 | **1** | 4 | **1** | 13 | 12 | **1** | 13 | 12 | 7.4 |
| SVR-L | 18 | 18 | 18 | 18 | 18 | 18 | 18 | 18 | 18 | 18 | 18 | 18 | 18 | 18 | 18 | 18 |
| SVR-R | 17 | 17 | 17 | 17 | 17 | 17 | 17 | 17 | 17 | 17 | 17 | 17 | 17 | 17 | 17 | 17 |
| XGB | 13 | 13 | 6 | 1 | 12 | 14 | 9 | 4 | 3 | 4 | 6 | 10 | 4 | 6 | 10 | 8.3 |

### 4.2.3 Algorithm Performance Summary

Polynomial Regression emerged as the most robust algorithm across the (0.0–1.0) and (0.9–1.0) reliability ranges, demonstrating superior consistency and ranking. Notably, in the (0.99–1.0) range, the difference between the best-performing algorithm and Polynomial Regression in terms of "Test MSE" did not exceed $9.14 \times 10^{-5}$ ("Test MSE gap" column in Table 7). These findings underscore PR's adaptability to varying reliability conditions, positioning it as a preferred choice for practical reliability estimation tasks even in the (0.99–1.0) range.

**Table 7.** Dominance of Polynomial Regression (PR) in the ultra-high-reliability range (0.99–1.0): rankings across benchmark networks.

| ID | Best Algorithm | Test MSE | Polynomial Regression | | |
|---|---|---|---|---|---|
| | | | Ranking | Test MSE | Test MSE gap |
| 1 | Polynomial Regression | 6.32673E-05 | 1 | 6.32673E-05 | 0.00E+00 |
| 2 | Linear Regression | 7.98783E-05 | 4 | 8.26034E-05 | 2.73E-06 |
| 3 | Bayesian Ridge | 9.08375E-07 | 15 | 1.56999E-06 | 6.62E-07 |
| 4 | Huber Regressor | 4.86874E-05 | 16 | 9.8913E-05 | 5.02E-05 |
| 5 | Bayesian Ridge | 6.71619E-05 | 8 | 6.77556E-05 | 5.94E-07 |
| 6 | Bayesian Ridge | 1.93685E-06 | 9 | 2.99513E-06 | 1.06E-06 |
| 7 | XGBoost | 6.55651E-07 | 13 | 3.56674E-06 | 2.91E-06 |
| 8 | Huber Regressor | 4.74994E-05 | 15 | 6.1596E-05 | 1.41E-05 |
| 9 | Extra Trees | 5.35769E-05 | 8 | 5.39464E-05 | 3.69E-07 |
| 10 | CatBoost | 5.9014E-05 | 13 | 6.31675E-05 | 4.15E-06 |
| 11 | Ridge Regression | 1.46747E-06 | 14 | 7.2372E-06 | 5.77E-06 |
| 12 | Lasso | 9.72152E-07 | 16 | 6.81341E-06 | 5.84E-06 |
| 13 | Ridge Regression | 7.36117E-07 | 12 | 3.85463E-06 | 3.12E-06 |
| 14 | Bayesian Ridge | 0.000244143 | 16 | 0.000333124 | 8.90E-05 |
| 15 | Bayesian Ridge | 0.000104019 | 16 | 0.000195459 | **9.14E-05** |

## 4.3 Experiment 2 (Ex 2)

Expanding upon Experiment 1, this phase evaluates 20 computational algorithms using the larger DM-LC biological network (39 nodes, 170 arcs) [37] as shown in Figure 7 across three reliability intervals: [0.0, 1.0], [0.9, 1.0], and [0.99, 1.0], starting with 10,000 samples.

### 4.3.1 Test Networks and their information

The DM-LC network serves as our scalability analysis testbed. While we initially defined three reliability intervals (full range: 0.0–1.0, high-reliability: 0.9–1.0, and ultra-high-reliability: 0.99–1.0), we discovered that all 10,000 samples in both high-reliability intervals yielded network reliability values of 1.0 (Table 8). Table 8 details statistical summaries of Test MSE, including the number of instances with approximated reliability equal to one ($N_1$), mean (Avg), standard deviation (Stdev), median (Med), minimum (Min), maximum (Max), and range (Range).

This occurred because the network's complexity prevents disconnections between nodes 1 and *n* under high component reliability conditions, making these intervals computationally trivial.

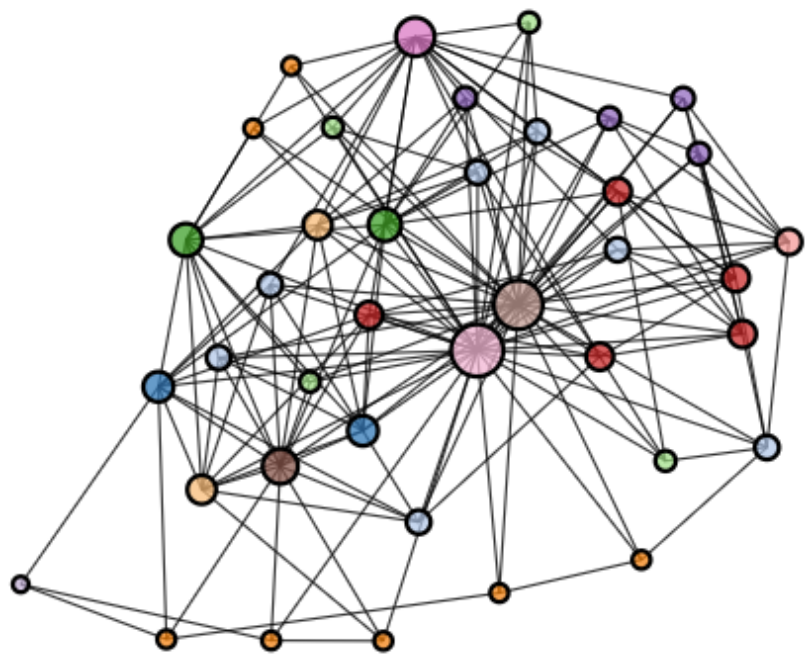

**Figure 7.** DM-LC.

We therefore redirected our focus to dataset scalability within the full reliability range (0.0–1.0), generating four progressively larger datasets: 10,000, 20,000, 30,000, and 40,000 samples. Each subsequent dataset incorporated all prior samples plus new records, ensuring controlled complexity escalation. All datasets were synthesized using the BAT-MCS framework.

**Table 8.** Statistical summary of approximated reliability instances in Ex2.

| Reliability Range | Size | $N_1$ | Avg | Stdev | Min | Max | Range | Med |
|---|---|---|---|---|---|---|---|---|
| (0.0–1.0)-reliability | 10,000 | 0 | 0.905862 | 0.049693 | 0.5694 | 0.9853 | 0.4159 | 0.9160 |
| | 20,000 | 0 | 0.905795 | 0.049459 | 0.5427 | 0.9874 | 0.4447 | 0.9158 |
| | 30,000 | 0 | 0.905892 | 0.049585 | 0.5427 | 0.9875 | 0.4448 | 0.9161 |
| | 40,000 | 0 | 0.905898 | 0.049361 | 0.5413 | 0.9895 | 0.4482 | 0.9162 |
| (0.9–1.0)-reliability | 10,000 | 10,000 | 1 | 0 | 1 | 1 | 0 | 1 |
| (0.99–1.0)-reliability | 10,000 | 10,000 | 1 | 0 | 1 | 1 | 0 | 1 |

### 4.3.2 The Observations

Algorithm performance exhibits strong dependence on dataset scale, as evidenced by diverging trends among methods (Table 9). Neural Networks (ANN) dominate smaller datasets (10,000–30,000 samples), achieving the lowest Test-MSE (e.g., 5.66E-05 at 40,000 samples), but relinquish their top rank to Polynomial Regression (PR) at 40,000 samples. PR's performance improves dramatically with data volume, rising from rank 6 (10,000 samples) to rank 1 (40,000 samples), where it marginally outperforms ANN (Test-MSE: 5.61E-05 vs. 5.66E-05). This shift reflects PR's quadratic parameter

complexity—requiring $m(m+1)/2$ coefficients for degree-$m$ interactions—which demands dataset scaling proportional to $O(m^2)$ to stabilize estimation.

Linear/Ridge/Bayesian Regression models consistently underperform (Test-MSE > **0.000264**), while tree-based methods (AdaBoost, XGBoost) and kernel-based approaches (Kernel Ridge, SVR) occupy mid- to low-tier ranks, underscoring their limited suitability for reliability prediction. Notably, ANN and PR exhibit continued accuracy gains at 40,000 samples, with PR's Test-MSE improvement (**2.49E-05**) exceeding ANN's (**1.58E-05**), challenging assumptions of strict diminishing returns.

**Table 9.** Ranking of each algorithm tested in Ex2.

| | 10,000 | 20,000 | 30,000 | 40,000 |
|---|---|---|---|---|
| AdaBoost | 13 | 13 | 13 | 14 |
| Bayesian Ridge | 3 | 3 | 4 | 4 |
| CatBoost | 8 | 9 | 10 | 10 |
| Decision Tree | 14 | 14 | 14 | 13 |
| ElasticNet | 17 | 17 | 17 | 17 |
| Extra Trees | 11 | 11 | 11 | 11 |
| Gradient Boosting | 10 | 8 | 8 | 9 |
| Huber Regressor | 5 | 5 | 6 | 6 |
| Kernel Ridge | 20 | 20 | 20 | 20 |
| K-Nearest Neighbors | 15 | 15 | 15 | 15 |
| Lasso | 16 | 16 | 16 | 16 |
| LightGBM | 9 | 6 | 9 | 8 |
| Linear Regression | 4 | 2 | 3 | 3 |
| Neural Network (ANN) | **1** | **1** | **1** | 2 |
| Polynomial Regression | 6 | 10 | 2 | **1** |
| Random Forest | 12 | 12 | 12 | 12 |
| Ridge Regression | 2 | 4 | 5 | 5 |
| SVR (Linear kernel) | 18 | 18 | 18 | 18 |
| SVR (RBF kernel) | 19 | 19 | 19 | 19 |
| XGBoost | 7 | 7 | 7 | 7 |

### 4.3.3 Algorithm Selection Framework

The empirical trends inform a selection framework prioritizing dataset scale, computational resources, and interpretability needs. For datasets of 10,000–30,000 samples, ANN achieves optimal accuracy (Test-MSE: 7.24E-05 at 30,000 samples) by adaptively modeling nonlinear relationships. At larger scales (40,000+ samples), PR surpasses ANN in accuracy (5.61E-05) while offering mathematical transparency, albeit requiring rigorous data scaling to meet its parametric demands.

Computational constraints further refine choices: ANN balances accuracy and training efficiency for moderate datasets, whereas PR excels in data-rich environments. Interpretability requirements favor

PR for safety-critical applications, while ANN suits accuracy-driven tasks accepting "black-box" tradeoffs. Tree-based methods, though excluded from this study, hold promise for sparse-data scalability and warrant future integration.

In practice, early-stage research prioritizes ANN for exploratory accuracy with limited data, while industrial applications leverage PR's interpretability at scale. Real-time systems benefit from PR's rapid inference post-training. The framework remains dynamic, requiring adaptation to emerging data paradigms (e.g., streaming) and algorithmic advancements.

**Table 10.** Results for the top three algorithms tested in Ex2.

| ID | Model | Train MSE | Test MSE | Test MAE | $R^2$ Score | CV Score |
|---|---|---|---|---|---|---|
| 10,000 | Neural Network (ANN) | 8.65E-05 | 0.000176 | 0.009863 | 0.923098 | 0.000124 |
| | Ridge Regression | 0.000297 | 0.000264 | 0.011494 | 0.884529 | 0.000302 |
| | Bayesian Ridge | 0.000297 | 0.000264 | 0.011512 | 0.884434 | 0.000302 |
| 20,000 | Neural Network (ANN) | 4.66E-05 | 7.55E-05 | 0.006208 | 0.970566 | 9.57E-05 |
| | Linear Regression | 0.000279 | 0.000307 | 0.011636 | 0.880143 | 0.00029 |
| | Bayesian Ridge | 0.000279 | 0.000308 | 0.011628 | 0.880092 | 0.00029 |
| 30,000 | Neural Network (ANN) | 5.02E-05 | 7.24E-05 | 0.006098 | 0.972025 | 6.55E-05 |
| | Polynomial Regression | 1.13E-05 | 8.1E-05 | 0.006868 | 0.968704 | 7.81E-05 |
| | Linear Regression | 0.000281 | 0.000315 | 0.011885 | 0.878162 | 0.000291 |
| 40,000 | Polynomial Regression | 1.6E-05 | 5.61E-05 | 0.005628 | 0.977208 | 5.63E-05 |
| | Neural Network (ANN) | 4.63E-05 | 5.66E-05 | 0.005489 | 0.977004 | 5.73E-05 |
| | Linear Regression | 0.00028 | 0.000293 | 0.011689 | 0.88107 | 0.000285 |

## 5. CONCLUSIONS

This study systematically evaluated 20 machine learning methods for binary-state network reliability across three regimes: full range (0.0–1.0), high reliability (0.9–1.0), and ultra-high reliability (0.99–1.0). Two critical contributions emerge, advancing both theoretical and practical aspects of reliability engineering.

To our knowledge, this work is the first to identify (i) a reliability convergence threshold for high-reliability networks and (ii) a dataset-size-driven performance crossover between ANN and PR. First, we demonstrate that in large-scale networks with arc reliability ≥0.9, system reliability converges to unity. This finding renders exhaustive reliability calculations computationally redundant for such systems, emphasizing that enhancing individual component reliability outweighs structural optimizations in ensuring robustness. Second, we establish a dataset-scale-dependent paradigm for algorithm selection. Artificial Neural Networks (ANN) excel in data-scarce regimes (size < $m^2$, where $m$ is the number of arcs), effectively approximating reliability functions with limited training data. In contrast, Polynomial Regression (PR) achieves superior accuracy in data-rich environments (size ≥ $m^2$), as its parameter estimation stabilizes with $O(m^2)$ samples.

These insights offer actionable guidelines: practitioners may bypass exhaustive computations for high-reliability networks (≥0.9) and adopt ANN or PR based on data availability. While this study focuses on binary-state networks, real-world systems often exhibit multi-state behavior with components operating at degraded or variable capacities. Future work will extend this framework to multi-state reliability analysis, addressing time-dependent degradation and heterogeneous performance levels, thereby enhancing applicability to complex infrastructure and engineering systems.

## ACKNOWLEDGMENT

This research was supported in part by the Ministry of Science and Technology, R.O.C. under grant MOST 110-2221-E-007-107-MY3. This article was once submitted to arXiv as a temporary submission that was just for reference and did not provide the copyright.

## REFERENCES

[1] S Al-Dahidi, P Baraldi, M Fresc, E Zio, L Montelatici. Feature Selection by Binary Differential Evolution for Predicting the Energy Production of a Wind Plant. *Energies* 2024, 17 (10): 2424.

[2] WC Yeh. Enhancing Reliability Calculation for One-Output k-out-of-n Binary-state Networks Using a New BAT. *Reliability Engineering and System Safety* 2025, 10.1016/j.ress.2025.110835.

[3] YH Lin, YF Li, E Zio. A reliability assessment framework for systems with degradation dependency by combining binary decision diagrams and Monte Carlo simulation. *IEEE Transactions on Systems, Man, and Cybernetics: Systems* 2015, 46 (11): 1556-1564.

[4] TP Nguyen, YK Lin. Reliability of a Multiple-Demand Multistate Air Transport Network With Flight Delays and Budget Constraints. *IEEE Transactions on Reliability* 2023.

[5] YF Niu, MM Yuan, XZ Xu. Optimal carrier selection to improve logistics network reliability with delivery spoilage. *Annals of Operations Research* 2025, DOI:10.1007/s10479-025-06521-y

[6] TJ Hsieh. A Q-learning guided search for developing a hybrid of mixed redundancy strategies to improve system reliability. *Reliability Engineering & System Safety* 2023, 236: 109297.

[7] RT Khameneh, K Barker, JE Ramirez-Marquez. A hybrid machine learning and simulation framework for modeling and understanding disinformation-induced disruptions in public transit systems. *Reliability Engineering & System Safety* 2025, 255: 110656.

[8] Z Hao, WC Yeh. GE-MBAT: An Efficient Algorithm for Reliability Assessment in Multi-State Flow Networks. *Reliability Engineering & System Safety* 2025, 110916.

[9] WC Yeh. A New Hybrid Inequality BAT for Comprehensive All-Level d-MP Identification Using Minimal Paths in Multistate Flow Network Reliability Analysis. *Reliability Engineering & System Safety* 2024, 244: 109876.

[10] Y Ke, X Wang, Z Ye, S Zhang, Z Cai. Binary decision diagram-based reliability modeling of phased-mission manufacturing system processing multi-type products. *Quality Technology & Quantitative Management* 2024, 21 (6): 1058-1075.

[11] WC Yeh, W Zhu. Optimal Allocation of Financial Resources for Ensuring Reliable Resilience in Binary-State Network Infrastructure. *Reliability Engineering & System Safety* 2024, 250: 110265.

[12] E Zio, N Pedroni. Reliability Estimation by Advanced Monte Carlo Simulation. Editors: Faulin, Juan, Martorell, Ramirez-Marquez. *Simulation Methods for Reliability and Availability of Complex Systems* 2010, 3-39, Springer Series in Reliability Engineering.

[13] CF Huang. A Monte Carlo-based algorithm for the quickest path flow network reliability problem. *Annals of Operations Research* 2024, https://doi.org/10.1007/s10479-024-06377-8.

[14] WC Yeh. Novel Self-Adaptive Monte Carlo Simulation Based on Binary-Addition-Tree Algorithm for Binary-State Network Reliability Approximation. *Reliability Engineering & System Safety* 2022, 228: 108796.

[15] L Friedli, N Linde. Rare event probability estimation for groundwater inverse problems with a two-stage Sequential Monte Carlo approach. *Water Resources Research* 2024, https://doi.org/10.1029/2023WR036610

[16] WC Yeh. A Quick BAT for Evaluating the Reliability of Binary-State Networks. *Reliability Engineering & System Safety* 2021, 216: 107917.

[17] CJ Colbourn. The combinatorics of network reliability. Oxford University Press, Inc. 1987.

[18] DR Shier. Network reliability and algebraic structures. Clarendon Press, 1991.

[19] WC Yeh, CC Kuo. Predicting and modeling wildfire propagation areas with BAT and maximum-state PageRank. *Applied Sciences* 2020, 10 (23): 8349.

[20] WC Yeh. Novel Algorithm for Computing All-Pairs Homogeneity-Arc Binary-State Undirected Network Reliability. *Reliability Engineering & System Safety* 2021, 216: 107950.

[21] WC Yeh, SY Tan, W Zhu, CL Huang, G Yang. Novel Binary Addition Tree Algorithm (BAT) for Calculating the Direct Lower-Bound of the Highly Reliable Binary-State Network Reliability. *Reliability Engineering & System Safety* 2022, 223: 108509.

[22] WC Yeh, W Zhu, CL Huang, TY Hsu, Z Liu, SY Tan. A New BAT and PageRank Algorithm for Propagation Probability in Social Networks. *Applied Sciences* 2022, 12: doi.org/10.3390/app12146858

[23] W Xia, Y Wang, Y Hao, Z He, K Yan, F Zhao. Reliability analysis for complex electromechanical multi-state systems utilizing universal generating function techniques. *Reliability Engineering & System Safety* 2024, 244: 109911.

[24] WC Yeh, CM Du, SY Tan, M Forghani-elahabad. Application of LSTM Based on the BAT-MCS for Binary-State Network Approximated Time-Dependent Reliability Problems. *Reliability Engineering & System Safety* 2023, 235: 108954.

[25] H Dui, H Li, X Dong, S Wu. An energy IoT-driven multi-dimension resilience methodology of smart microgrids. *Reliability Engineering & System Safety* 2025, 253: 110533.

[26] M Zhu, X Huang, H Pham. A random-field-environment-based multidimensional time-dependent resilience modeling of complex systems. *IEEE Transactions on Computational Social Systems* 2021, 8 (6): 1427-1437.

[27] YZ Su, WC Yeh. The protection and recovery strategy development of dynamic resilience analysis and cost consideration in the infrastructure network. *Journal of Computational Design and Engineering* 2022, 9 (1): 168–186.

[28] WC Yeh. Time-reliability optimization for the stochastic traveling salesman problem. *Reliability Engineering & System Safety* 2024, 248: 110179.

[29] YZ Su, WC Yeh. Binary-addition tree algorithm-based resilience assessment for binary-state network problems. *Electronics* 2020, 9 (8): 1207.

[30] WC Yeh, E Lin, CL Huang. Predicting Spread Probability of Learning-Effect Computer Virus. *Complexity* 2021, 2021: 6672630.

[31] WC Yeh. BAT-based Algorithm for Finding All Pareto Solutions of the Series-Parallel Redundancy Allocation Problem with Mixed Components. *Reliability Engineering & System Safety* 2022, 228: 108795.

[32] WC Yeh. Novel Recursive Inclusion-Exclusion Technology Based on BAT and MPs for Heterogeneous-Arc Binary-State Network Reliability Problems. Reliability Engineering & System Safety 2022, 231: 108994.

[33] WC Yeh. A revised layered-network algorithm to search for all *d*-minpaths of a limited-flow acyclic network. *IEEE Transactions on Reliability* 1998, 47 (4): 436-442.

[34] WC Yeh. Novel binary-addition tree algorithm (BAT) for binary-state network reliability problem. *Reliability Engineering & System Safety* 2021, 208: 107448.

[35] Shier D, Network Reliability and Algebraic Structures, Clarendon Press, New York, NY, USA, 1991.

[36] Colbourn CJ, The combinatorics of network reliability, Oxford University Press, New York, 1987.

[37] https://networkrepository.com/bio-DM-LC.php